\documentclass[conference]{IEEEtran}
\IEEEoverridecommandlockouts
\usepackage{cite}
\usepackage{amsmath,amssymb,amsfonts}
\usepackage{algorithmic}
\usepackage{graphicx}
\usepackage{textcomp}
\usepackage{xcolor}

\usepackage{booktabs}
\usepackage{acronym}
\usepackage{dsfont}
\usepackage[inline]{enumitem}
\usepackage{hyperref}
\usepackage[utf8]{inputenc}
\usepackage[T1]{fontenc}

\def\BibTeX{{\rm B\kern-.05em{\sc i\kern-.025em b}\kern-.08em
    T\kern-.1667em\lower.7ex\hbox{E}\kern-.125emX}}

\DeclareMathOperator*{\argmin}{arg\,min \;}
\newcommand*{\name}[1]{\textsc{#1}}
\newcommand{\ie}{i.e.,~}
\newcommand{\eg}{e.g.,~}
\newcommand{\res}[2]{#1\(\,\pm\,\)#2}
\newcommand{\bres}[2]{\textbf{#1}\(\,\pm\,\)\textbf{#2}}
\newcommand{\sres}[2]{\(\underline{#1 \pm #2}\)}
\newcommand{\NewR}{\ensuremath{\mathds{R}}}

\newacro{AutoML}[AutoML]{automated machine learning}
\newacro{CNN}[CNN]{convolutional neural network}
\newacro{GRU}[GRU]{gated recurrent unit}
\newacro{LSTM}[LSTM]{long short-term memory}
\newacro{ML}[ML]{machine learning}
\newacro{NN}[NN]{neural network}
\newacro{PHM}[PHM]{prognostics and health management}
\newacro{RMSE}[RMSE]{root mean square error}
\newacro{RUL}[RUL]{remaining useful life}
\newacro{SME}[SME]{small and medium-sized enterprise}
\newacro{seq2seq}[seq2seq]{sequence-to-sequence}
\newacro{SVM}[SVM]{support vector machine}
\newacro{TCN}[TCN]{temporal convolutional network}
\newacro{EoL}[EoL]{end of life}
\newacro{TSA}[TSA]{time synchronous averaging}
\newacro{HPO}[HPO]{hyperparameter optimization}
\newacro{DCAE}[DCAE]{deep convolutional autoencoder}
\newacro{HI}[HI]{health indicator}
\newacro{HEdA}[HEA]{human expert approach}
\newacro{RF}[RF]{random forest}
\newacro{ES}[ES]{engineering system}

\begin{document}

\title{Automated Machine Learning for Remaining Useful Life Predictions\\
}

\author{\IEEEauthorblockN{Marc-Andr\'e Z\"oller\textsuperscript{\textsection}}
\IEEEauthorblockA{\textit{USU Software AG} \\
Karlsruhe, Germany \\
marc.zoeller@usu.com}
\and
\IEEEauthorblockN{Fabian Mauthe\textsuperscript{\textsection}}
\IEEEauthorblockA{\textit{Esslingen University of Applied Sciences}\\
Esslingen, Germany \\
fabian.mauthe@hs-esslingen.de}
\and
\IEEEauthorblockN{Peter Zeiler}
\IEEEauthorblockA{\textit{Esslingen University of Applied Sciences}\\
Esslingen, Germany \\
peter.zeiler@hs-esslingen.de}
\and
\IEEEauthorblockN{Marius Lindauer}
\IEEEauthorblockA{\textit{Institute of Artificial Intelligence} \\
\textit{Leibniz University Hannover}\\
Hannover, Germany \\
m.lindauer@ai.uni-hannover.de}
\and
\IEEEauthorblockN{Marco F. Huber}
\IEEEauthorblockA{\textit{Institute of Industrial Manufacturing and Management IFF, University of Stuttgart}\\
\textit{Department Cyber Cognitive Intelligent (CCI), Fraunhofer IPA}\\
Stuttgart, Germany \\
marco.huber@ieee.org}
}

\maketitle

\begingroup\renewcommand\thefootnote{\textsection}
\begin{NoHyper}
\footnotetext{These authors contributed equally}
\end{NoHyper}
\endgroup

\begin{abstract}
Being able to predict the \ac{RUL} of an \acl{ES} is an important task in \acl{PHM}. Recently, data-driven approaches to \ac{RUL} predictions are becoming prevalent over model-based approaches since no underlying physical knowledge of the \acl{ES} is required. Yet, this just replaces required expertise of the underlying physics with \ac{ML} expertise, which is often also not available. \Ac{AutoML} promises to build end-to-end \ac{ML} pipelines automatically enabling domain experts without \ac{ML} expertise to create their own models. This paper introduces \name{AutoRUL}, an \ac{AutoML}-driven end-to-end approach for automatic \ac{RUL} predictions. \name{AutoRUL} combines fine-tuned standard regression methods to an ensemble with high predictive power. By evaluating the proposed method on eight real-world and synthetic datasets against state-of-the-art hand-crafted models, we show that \ac{AutoML} provides a viable alternative to hand-crafted data-driven \ac{RUL} predictions. Consequently, creating \ac{RUL} predictions can be made more accessible for domain experts using \ac{AutoML} by eliminating \ac{ML} expertise from data-driven model construction.
\end{abstract}


\section{Introduction}

In recent manufacturing, a reliable, available, and sustainable production of goods is important to be competitive. This requires advanced maintenance strategies such as predictive maintenance. Traditional strategies such as corrective or preventive maintenance cause unplanned downtime or do not utilize existing resources completely due to superfluous maintenance actions. Predictive maintenance can help to avoid unplanned downtime while also reducing unnecessary maintenance costs. Knowledge about the future degradation behavior of an \ac{ES} is crucial to plan the required maintenance as predictive maintenance is becoming more important in industry. The engineering discipline of \acf{PHM} studies techniques for transitioning from corrective or preventive maintenance to predictive maintenance. A key task in \ac{PHM} is the prognosis of the \acf{RUL}. Approaches are often divided into model-based, data-driven, and hybrid methods \cite{Atamuradov.2017}. Usually, the \ac{RUL} is used to plan the next maintenance and has attracted considerable interest in the research community.

Model-based approaches use mathematical descriptions, like algebraic and differential equations or physics-based models to predict the future degradation behavior of a system \cite{Zio.2022}. Such models require thorough understanding of the mechanism involved in the degradation process and are time-consuming to create. Also, not all systems can be expressed with sufficient precision in solvable mathematical or physics models due to the complexity of real world systems \cite{Kordestani.2021}. With the integration of more sensors, the amount of available data about the condition of a system is gradually increasing. In parallel, \acf{ML} has matured and achieved wide-spread dissemination. Similarly, data-driven models for \ac{RUL} have become more popular. By using \ac{ML} to model the relationship of the system health and \ac{RUL}, no thorough understanding of the degradation process is required. In recent years, various \ac{ML} methods, like \acl{RF} or different \aclp{NN}, have been proposed to model \ac{RUL} predictions \cite{Zio.2022, Mauthe2022}.

Yet, to actually create data-driven models for \ac{RUL} predictions from recorded sensor data, specialized knowledge in data modeling and \ac{ML} is necessary. For \aclp{SME} such knowledge is often not available. As a consequence, the potential of data-driven methods is rarely utilized. \Acf{AutoML} promises to enable domain experts, \ie the manufacturer of an \ac{ES}, to develop data-driven \ac{ML} methods for \ac{RUL} prediction without detailed knowledge of \ac{ML} \cite{Hutter2018}. Enterprises with \ac{ML} knowledge can benefit from \ac{AutoML} by automating time-consuming \ac{ML} sub-tasks, leading to lower costs and higher competitiveness.

The contributions of this paper are as follows:
\begin{enumerate}
    \item We propose a universal end-to-end \ac{AutoML} solution, called \name{AutoRUL}, for \ac{RUL} predictions for domain experts that does not require \ac{ML} knowledge.
    \item We compare the \textit{traditional} manual creation of \ac{RUL} predictions with the proposed automatic approach. 
    \item \name{AutoRUL} is validated against five state-of-the-art \ac{RUL} models on a wide range of real-world and synthetic datasets. This is the first evaluation of a single \ac{AutoML} tool on a wide variety of \ac{RUL} datasets.
    \item In the spirit of reproducible research, all source code and used datasets are publicly available.
\end{enumerate}

Section~\ref{sec:related_work} introduces related work. Our proposed approach is introduced in Section~\ref{sec:rul_predictions} and compared in Section~\ref{sec:comparison}. In Section~\ref{sec:experiments} the proposed method is validated on \ac{RUL} datasets. The paper concludes in a short discussion in Section~\ref{sec:conclusion}.

\section{Related Work}
\label{sec:related_work}

\subsection{Remaining Useful Life Prognosis}
\label{sec:rul}

Key tasks in \ac{PHM} are the fault detection, diagnosing which component of an \ac{ES} causes a present fault condition, the health assessment, and the prognosis of the \ac{RUL} of a system \cite{Zio.2022}. 
Each prognosis approach requires---or at least benefits from---the availability of sufficient condition data of the system. The knowledge of future degradation behavior and \ac{RUL} is crucial for maintenance strategies such as predictive maintenance.

\ac{RUL} prediction can be described as estimating the remaining time duration that an \ac{ES} is likely to continue to operate before maintenance is required or a failure occurs \cite{Kordestani.2021}. There are two main strategies, indirect and direct, for predicting \ac{RUL} using a data-driven model \cite{Goebel.2008}. Indirect \ac{RUL} prognosis first predicts future degradation behavior. At the time of prediction, the model performs an extrapolation based on present condition data until a failure criterion is achieved. The time span of this extrapolation corresponds to the actual predicted \ac{RUL}. In contrast, in the direct approach, the \ac{RUL} is predicted directly given the current condition data. This requires models determining the correlation between condition data and \ac{RUL} based on a defined failure criterion. In this paper, we consider a direct data-driven \ac{RUL} prognosis. 

More formally, data-driven \ac{RUL} predictions is formulated as a sequence-to-target learning problem aiming to learn a mapping \(h: \mathcal{X} \rightarrow \mathcal{Y}\) from historical condition data \(\vec{x} \in \mathcal{X}\)---recorded over various instances, \ie \acp{ES}---to the \ac{RUL} \(y \in \mathcal{Y}\subset \NewR\). Each instance \(\vec{x}_i\) contains a \(d = |\vec{x}_i|\) dimensional (multivariate) time series of arbitrary length with sensor data from the system setup until the defined failure criterion. Training data is defined as \(\mathcal{D}_\mathrm{train} = \{(\vec{x}_i, y_i)\}_{i = 1}^N\) with \(N\) being the number of instances. Test data \(\mathcal{D}_\mathrm{test} = \{(\vec{x}'_i, y'_i)\}_{i = 1}^{M} \) contain sensor data up to some time \(T_i\) with \(y_i\) being the corresponding \ac{RUL} that is left after \(T_i\). A loss function \(\mathcal{L}\) is used to measure the performance of \(h\) on \(\mathcal{D}_\mathrm{test}\). In the context of \ac{RUL}, loss functions can be divided into general symmetric loss functions or specialized asymmetric loss functions. 
We use the \ac{RMSE}
\begin{equation}
\label{eq:rmse}
    \mathcal{L}_\mathrm{RMSE} \left(h, \mathcal{D}_\mathrm{test}\right) := \sqrt{ \dfrac{1}{M} \sum_{i = 1}^M \bigl( h(\vec{x}'_i) - y'_i \bigr)^2}
\end{equation}
with \(h(\vec{x}'_i)\) being the predicted \ac{RUL} on \(\vec{x}'_i\) in this work, but any other loss function for regression would also be applicable.

\subsection{Human Expert Approach}
\label{sec:human_expert}
The elementary phases in developing an \ac{ML} model for \ac{RUL} predictions are similar to developing a model for regressions, namely \textit{data pre-processing}, \textit{feature engineering}, and the \textit{\ac{ML} model development}. Yet, expert knowledge regarding \ac{RUL} prognosis as well as the \ac{ES} can be used to integrate additional information into the respective phases \cite{Kordestani.2021}.

\subsubsection{Data Pre-Processing}
Basis for prediction models are historical raw signals recorded by various sensors of the respective \acp{ES}. Typically, these signals contain shortcomings, \eg noise \cite{Zio.2022,Atamuradov.2017}. Therefore, signal processing methods, like smoothing or imputation, are used to increase the data quality. In addition, different methods can be applied to subsequently generate informative features that enable the \ac{RUL} to be mapped, \eg the generation of a frequency spectrum \cite{JundaZhu.2014}. 

\subsubsection{Feature Engineering}
Next, new meaningful features from the pre-processed signals are generated \cite{Atamuradov.2017}. This is crucial for the later modelling. Often, new features are generated on shorter \textit{time windows} containing only a few measurements. They can be grouped into \textit{common}---like statistical time-domain features---, \textit{specific}---developed specifically for an individual \ac{ES}---and \textit{\ac{ML}-based}---based on anomaly detection or one-class classification \cite{Wang.2018}---\aclp{HI} \cite{JundaZhu.2014,Wang.2018}. In addition, features taking operating conditions, \eg loading, into account are also essential. Finally, relevant features must be selected considering the correlation of the features with the target \ac{RUL}. Statistical hypothesis tests are often used for this \cite{Christ2016}. In case of too many relevant features, methods like autoencoders or PCA are used for dimension reduction \cite{Lu.2019}.

\begin{figure*}[t!]
    \centering
    \includegraphics[width=0.92\textwidth]{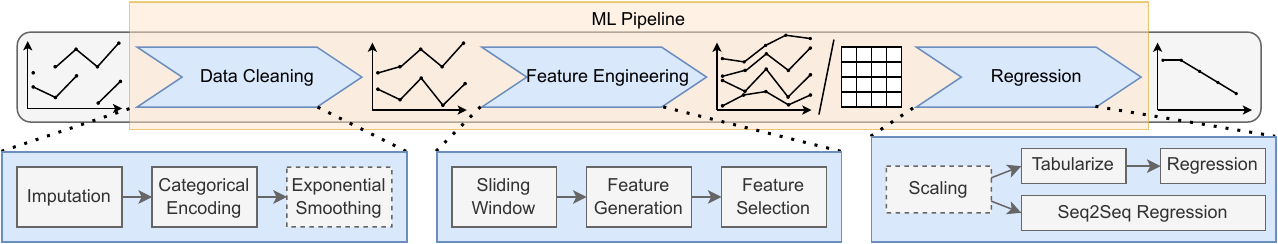}
    \caption{Schematic representation of pipelines constructed by \name{AutoRUL} to map sensor data to \ac{RUL} predictions. The pipeline is structured into three major phases: \textit{data cleaning}, \textit{feature engineering} and \textit{regression}. Each phase contains multiple steps, displayed below the \ac{ML} pipeline, configured by various hyperparameters. Steps with solid borders indicate mandatory steps included in every pipeline while dashed borders represent optional pipeline steps.}
    \label{fig:automl_procedure}
\end{figure*}

\subsubsection{ML Model Development}
Selecting the best \ac{ML} method for the \ac{ES} or condition data is a challenging task and requires sufficient \ac{ML} expertise. Various \ac{ML} methods like \ac{SVM} \cite{GarciaNieto.2015}, \ac{RF} \cite{Kuersat2020}, and different \acp{NN} are therefore available \cite{Zio.2022,Mauthe2022}. Versatile adaptations of such methods to the respective \acp{ES} or requirements can be found in the literature \cite{Zio.2022,Kordestani.2021}. \acp{NN}, for example, allow many adaptation options to the system through different network architectures. Recently, several specialized network architectures have been proposed \cite{Zio.2022,Kordestani.2021}. Mo et al. \cite{Mo2021a} proposed a transformer encoder-based model with an improved sensitivity for local contexts (time steps). Shi and Chehade \cite{Shi.2021} introduced a dual \ac{LSTM} framework for combining the detection of health states with the \ac{RUL} prediction for achieving better performance. Wang et al. \cite{Wang.2021} proposed a combination of an autoencoder for feature extraction and a \ac{CNN} for \ac{RUL} prediction.

This brief overview illustrates that numerous adaptations of a selected \ac{ML} method are possible. Especially the integration of additional expert knowledge about the \ac{ES} allows improving the prediction performance~\cite{Braig.2023,Hagmeyer.2022}.

\subsection{Automated Machine Learning}
\ac{AutoML} aims to construct \ac{ML} pipelines automatically with minimal human interaction \cite{Hutter2018}. Many different \ac{AutoML} approaches have been proposed with most propositions focusing on standard learning tasks, \eg tabular or image classification. Virtually all \ac{AutoML} approaches model the synthesis of \ac{ML} pipelines as a black-box optimization problem: Given a dataset and loss function, \ac{AutoML} searches for a pipeline in a predefined search space \(\Lambda\) minimizing the validation loss.

In general, various \ac{AutoML} tools for standard regression and time-series forecasting tasks exist. Yet, nearly all of these tools are of limited use for \ac{RUL} predictions due to two limitations:
\begin{enumerate*}
    \item Tools for standard regression assume tabular data and do not provide measures to transform time-series data into a tabular format.
    \item Tools for time-series forecasting train local statistical models that can forecast the future behavior of a single \ac{ES} without learning generalized concepts over multiple system instances.
\end{enumerate*}

More recently, dedicated \ac{AutoML} systems for \ac{RUL} predictions have been proposed. \name{ML-Plan-RUL} \cite{Tornede2020} was the first \ac{AutoML} tool tailored to \ac{RUL} predictions. Using descriptive statistics, the time series data is transformed to fixed, tabular data suitable for standard regression. In a follow-up work, the search space of \name{ML-Plan-RUL} is split into a feature engineering and regression part. By using a joint genetic algorithm over both search spaces, the improved approach, named \name{AutoCoevoRUL} \cite{Tornede2021}, was able to outperform \name{ML-Plan-RUL}. Similarly, Singh et al.~\cite{Singh2022} used a genetic algorithm to optimize the parameters of an \ac{LSTM} to predict the \ac{RUL} of batteries. Finally, Zhou et al.~\cite{Zhou2020} used a combination of \ac{LSTM} and \ac{CNN} cells for \ac{RUL} predictions with the exact network architecture being determined using \ac{AutoML}.

\section{Automated Remaining Useful Life Predictions }
\label{sec:rul_predictions}
\ac{AutoML} frameworks usually consider a combined search space covering all necessary steps to assemble an \ac{ML} pipeline for a specific task. The search space covers both the selection of specific algorithms, \eg using an \ac{SVM} for regression, as well as the configuration of the selected algorithms via hyperparameters. Considering the task of \ac{RUL} prediction, a complete \ac{ML} pipeline has to be able to consume the raw sensor data as univariate or multivariate time series, pre-process the data and finally use a regression algorithm to create \ac{RUL} predictions. In contrast to standard regression tasks, which assume tabular input data, the pipeline either has to include methods to transform time series into tabular data, or the regression algorithm has to be capable of consuming time series data directly, like, for example, \acp{LSTM}.

Our approach, \name{AutoRUL}, uses a flexible best-practice pipeline, inspired by human experts, displayed in Fig.~\ref{fig:automl_procedure}. It starts with basic data cleaning, followed by a feature generation step and an optional transformation into tabular data if necessary. A regressor is used to predict the \ac{RUL}. \ac{AutoML} aims to fill these generic pipeline steps with actual algorithms, configured by respective hyperparameters, such that the loss function~\eqref{eq:rmse} is minimized on a validation dataset \(\mathcal{D}_{\mathrm{val}}\). The according optimization problem is formulated as
\begin{equation}
\label{eq:automl_optimization}
	\vec{\lambda}^* \in \argmin_{\vec{\lambda} \in \Lambda} \mathcal{L}_\mathrm{RMSE} \left(h_{\vec{\lambda}}, \mathcal{D}_\mathrm{val} \right)
\end{equation}
with \(h_{\vec{\lambda}}\) being the pipeline defined by the \textit{configuration} \(\vec{\lambda}\), \ie a set of selected algorithms and hyperparameters. To actually solve \eqref{eq:automl_optimization}, we use Bayesian optimization, namely \name{SMAC} \cite{Lindauer2022}. Bayesian optimization is an iterative optimization procedure for expensive objective functions. By repeatedly sampling pipeline candidates, an internal surrogate model of the objective function is constructed. This surrogate model is used to trade-off exploration of under-explored settings and exploitation of well-performing regions.

The search space \(\Lambda\) of \name{AutoRUL} allows the creation of \(624\) unique pipelines which can be further configured by a total of \(168\) hyperparameters. In the remainder of this section, further details regarding the search space definition as well as implemented performance improvements are presented. Further information are available in the online appendix.\footnote{
    \url{https://github.com/Ennosigaeon/auto-sktime/blob/main/smc/appendix.pdf}
}

\paragraph{Data Cleaning}
The first part of the generated \ac{ML} pipeline is dedicated to data cleaning to solve common data issues. At first, missing values for individual timestamps are always imputed, with the actual imputation method, \eg repeating neighbouring values or using the sequence mean value, being selected by the optimizer. Next, categorical variables are encoded to ensure compatibility with all later steps. Finally, an optional exponential smoothing of the data can be applied.

\paragraph{Feature Engineering}
Next, the raw sensor measurements are enriched with new features. Using a sliding window with tunable length \(\lambda_w\), a context with prior observations is created for each timestamp \(t\). Especially for high-frequent measurements, not every recording is relevant but increases the computational load. A configurable stride hyperparameter \(\lambda_s\) is used to determine the number of steps the window is shifted, effectively down-sampling \(\vec{x}_{i}\). By design \(\lambda_s\) is always smaller than \(\lambda_w\) leading to an overlap of adjacent windows.

In general, two different approaches are used to create new features from the generated windows:
\begin{enumerate*}
    \item The \(d \times \lambda_w\) measurements in a single window can be \textit{flattened} to a \(1 \times (d \cdot \lambda_w)\) vector to embed lagged observations.
    \item Each generated window can be interpreted as new time series with fixed length \(\lambda_w\). The new time series can be characterized using different statistical and stochastic features. \name{AutoRUL} supports \(43\) unique features and the optimizer can select which features to include.
\end{enumerate*}
In both cases the time series characteristics of the input data is preserved. During feature selection, irrelevant features are identified using statistical hypothesis tests to eliminate features uncorrelated with the regression target \cite{Christ2016}.

\paragraph{Regression}
The actual \ac{RUL} predictions are created by mapping the problem to a regression learning task. In contrast to other \ac{RUL} publications, we do not aim to propose a novel method for \ac{RUL} predictions, but instead focus on configuring and combining standard \ac{ML} regression models to achieve a good performance on a wide variety of datasets.

Recently, various \ac{seq2seq} models that consume the complete input sequence and produce a new sequence, \ie the \ac{RUL}, have been proposed for \ac{RUL} predictions. \name{AutoRUL} supports recurrent architectures, namely \ac{LSTM} and \ac{GRU}, and convolutional architectures, namely \ac{CNN} and \ac{TCN}. For each basis architecture, the concrete network architecture, like the size of latent spaces or number of layers, is configured via hyperparameters. In addition, the complete training process of the \ac{NN}, \eg the batch size or learning rate, is tunable via various hyperparameters.

Alternatively, \name{AutoRUL} supports splitting \(\vec{x}_i\) into chunks with fixed length, effectively transforming variable length input data into a tabular format with fixed dimensions. This allows employing traditional regression methods like \acp{RF}.

\paragraph{Performance Improvements}
Even though Bayesian optimization is quite efficient in the number of sampled configurations, \ac{AutoML} still evaluates a large number of potential solutions. Therefore, we implement four additional performance improvements. These improvements either aim to increase the performance of the final model or reduce the optimization duration:
\begin{enumerate*}
    \item State-of-the-art models for data-driven \ac{RUL} predictions, like \acp{NN}, are expensive to train. This problem becomes even more pressing for \ac{AutoML} as many different models have to be evaluated during the optimization. To compensate for this overhead, we employ a \textit{multi-fidelity} optimization strategy: By limiting the search space to models that support iterative fitting, models are only fitted for a few iterations and bad-performing models are discarded early while all remaining models are assigned more iterations. In the context of \name{AutoRUL}, iterations can, for example, represent the epochs of an \ac{NN} or number of trees in an \ac{RF}. The exact procedure how to allocate the number of iterations is handled by \name{Hyperband} \cite{Li2018}.
    \item In general, it is not possible to exactly predict how long a certain configuration will take to be fitted. Therefore, it may be possible that some sampled configurations require an unreasonable portion of the total optimization budget to evaluate. Consequently, less models will be evaluated in total limiting the potential of Bayesian optimization. To reduce the negative impact of these unpropitious configurations, the training of models is aborted after a user-provided timeout.
    \item During the optimization, \name{AutoRUL} creates multiple models. Instead of only using the best model and dropping all others, we create an ensemble of the best forecasters using ensemble selection \cite{Caruana2004}.
    \item Finally, as different candidate models are independent of each other, \name{AutoRUL} supports fitting models in parallel to achieve a better utilization of the available computing hardware.
\end{enumerate*}

\begin{table*}[t]
    \caption{Test \ac{RMSE} of \name{AutoRUL} and reproductions of state-of-the-art \ac{RUL} methods. Reported are mean and standard deviation over ten repetitions. The best results are highlighted in bold. Approaches that are not significantly worse  (Wilcoxon signed-ranked test) than the best result are highlighted with underscores. * indicates that the model was originally proposed for this dataset.}
    \label{tab:experiment-results}

    \setlength{\tabcolsep}{5pt}
    \centering
    \begin{tabular}{@{} l |c c c c c c c c @{}}
        \toprule
        Method                          & FD001                 & FD002                 & FD003                 & FD004                 & PHM'08                & PHME'20               & Filtration        &  PRONOSTIA           \\
        \midrule
        LSTM-FNN \cite{Zheng2017}       & \res{14.78}{0.57}*    & \res{17.98}{0.74}*    & \res{11.76}{0.33}*    & \bres{11.66}{0.33}*   & \sres{28.32}{0.88}*   & \res{16.82}{4.16}~    & ~\res{6.32}{0.71} & \res{35.58}{4.77}     \\ 
        CNN-FNN \cite{Li2018b}          & \res{16.86}{0.68}*    & \res{23.57}{0.33}*    & \res{13.79}{0.42}*    & \res{18.77}{0.71}*    & \res{31.47}{0.31}~    & \res{23.57}{3.57}~    & \res{15.00}{0.87} & \res{32.13}{2.63}     \\ 
        Transformer  \cite{Mo2021a}     & \res{13.85}{0.66}*    & \bres{14.46}{0.18}*   & \res{12.86}{0.12}*    & \res{12.50}{0.21}*    & \res{29.92}{0.54}~    & \sres{11.09}{1.78}~   & ~\res{6.97}{0.44} & \res{51.05}{1.74}     \\ 
        Random Forest \cite{Kuersat2020}& \res{14.39}{0.03}~    & \res{18.80}{0.05}~    & \res{12.41}{0.03}~    & \res{13.67}{0.03}~    & \res{31.01}{0.36}~    & \res{15.71}{4.49}*    & ~\res{6.69}{0.71} & \sres{27.29}{2.21}    \\
        SVM \cite{GarciaNieto.2015}     & \res{13.93}{0.28}*    & \res{20.18}{0.47}*    & \res{12.66}{0.31}*    & \res{16.12}{0.48}*    & \res{40.66}{6.26}~    & \res{29.47}{2.70}~    & ~\res{9.00}{0.73} & \res{27.90}{1.53}     \\ 
        AutoCoevoRUL \cite{Tornede2021} & \res{15.61}{0.76}*    & \res{17.79}{1.07}*    & \res{15.62}{0.88}*    & \res{16.34}{0.91}*    & \res{46.30}{1.15}*    & \res{15.23}{2.49}~    & \res{10.99}{0.38}   & \res{33.07}{5.87}   \\
        \name{AutoRUL}                  & \bres{9.21}{0.58}     & \sres{14.55}{0.45}~   & \bres{8.14}{0.39}     & \res{12.58}{0.15}~    & \bres{27.76}{1.19}~   & \bres{8.93}{4.94}     & ~\bres{5.85}{0.83}& \bres{22.52}{5.68}    \\
        \bottomrule
    \end{tabular}
\end{table*}

\section{Comparison between Approaches}
\label{sec:comparison}
The phases, visualized in Fig.~\ref{fig:automl_procedure}, of the proposed \name{AutoRUL} are generally identical to the phases of the \acf{HEdA} described in Section~\ref{sec:human_expert}. However, there are still some deviations:
\begin{enumerate*}
    \item In the first phase, the focus is on data cleaning and improving the data quality to enable the \ac{RUL} to be mapped. \name{AutoRUL} contains the same methods (imputation, encoding, smoothing) that are commonly used in the \ac{HEdA}. For individual \acfp{ES}, it is beneficial to apply different methods to generate additional signals, like the frequency spectrum. Thereby the subsequent feature generation can be improved. This is an individual step not included in \name{AutoRUL}. At this point, expert knowledge related to the \ac{ES} can be taken into account and specific properties can be considered. The data pre-processing in \name{AutoRUL} is limited to general issues that can occur in datasets. It is not possible to adapt the pre-processing to specific and known problems in the concrete dataset.
    \item For feature engineering, the steps time window processing, feature generation, and feature selection, are also included in \name{AutoRUL}. In \acp{HEdA}, however, feature generation is often much more extensive. \name{AutoRUL} contains \(43\) unique features that also cover some common \acfp{HI}, however no specific \acp{HI} or \ac{ML}-based \acp{HI} are possible to generate. In \name{AutoRUL} feature selection is solely performed by correlation of the features with the \ac{RUL}. However, for \acp{ES} it may happen that features are valuable despite a weak correlation, \eg load information. In \ac{HEdA}, these features can still be considered, whereas in \name{AutoRUL} the valuable information is lost.
    \item In the regression phase, \name{AutoRUL} contains many \ac{ML} methods that are also commonly used in the \ac{HEdA}. These are implemented in their basic form without special adaptations of the algorithms. In \ac{HEdA}, specific adaptations of the methods are often found to address characteristics of the \ac{ES}. For example, regarding the system, diagnoses can be combined with predictions. Only when a specific system condition is detected, the prognosis of the \ac{RUL} is carried out. 
\end{enumerate*}

Due to the mentioned deviations from \name{AutoRUL}, no specific characteristics of the \ac{ES} can be taken into account. Consequently, no knowledge about the system or the underlying degradation process is included and it is not possible to improve the model even though this knowledge is available. In turn, \name{AutoRUL} is easy to apply to different \acp{ES} with minimal human setup time. In contrast, \ac{HEdA} requires highly qualified human experts for developing the models. This includes proven \ac{ML} knowledge as well as technical expertise regarding the \ac{ES} to be able to implement an \ac{HEdA} successfully. This is often an obstacle for small and medium-sized enterprises, as it involves high resources for human experts. In addition, the \ac{HEdA} requires a high development~effort.

\section{Empirical Evaluation}
\label{sec:experiments}

To proof the viability of our proposed approach, we test \name{AutoRUL} on a wide variety of well-established \ac{RUL} datasets. As baseline measures, we try to replicate the results of state-of-the-art manual \ac{RUL} predictions and \name{AutoCoevoRUL}.\footnote{
    Unfortunately, replicating previous results exactly is often not possible as either no implementations were provided or crucial implementation details were not reported. We tried to replicate the results to the best of our abilities.
}

\paragraph{Evaluated Datasets}
The C-MAPSS dataset \cite{Saxena2008}---the most popular \ac{RUL} benchmark dataset---contains simulated run-to-failure sensor data of turbofan engines. It contains four different datasets, called \textit{FD001} to \textit{FD004}, that use different numbers of operating conditions and failure modes. Similarly, a fifth dataset of C-MAPSS was used in the PHM'08 data challenge. The PHME'20 data challenge \cite{Giordano2020} dataset contains real-world run-to-failure sensor data of a fluid filtration system. In multiple experiments filters are loaded with different particle sizes until they are clogged. The \name{PRONOSTIA} bearings dataset \cite{Nectoux2012} contains real-world degradation data of bearings at varying operation conditions. The bearings have reached their end-of-life when exceeding a predefined vibration threshold. The Filtration dataset \cite{Mauthe2022} contains real degradation data of air filters under different loading levels. Increasing differential pressure due to progressing filter clogging represents the degradation process. All datasets have a predefined holdout test set for the final performance evaluation. It is important to note that none of the data sets used for the empirical evaluation were used during development. Therefore, \name{AutoRUL} does not contain any optimizations for the evaluated datasets.

\paragraph{Experiment Setup}
All experiments were executed on virtual machines with four cores, 15 GB RAM and an NVIDIA T4 GPU. To account for non-determinism, ten repetitions with different initial seeds are used. Each pipeline candidate was trained for up to five minutes with a 80/20\% train/validation split. New candidates are sampled until the defined total budget of ten hours is exhausted. In addition, ensembles with at most ten different pipelines are constructed on the fly. To ensure reproducibility of all presented results, the source code, including scripts for the experiments, is available on \name{Github}.\footnote{
    See \url{https://github.com/Ennosigaeon/auto-sktime}.
}

\begin{figure}[t]
    \centering
    \centerline{\includegraphics[width=0.45\textwidth]{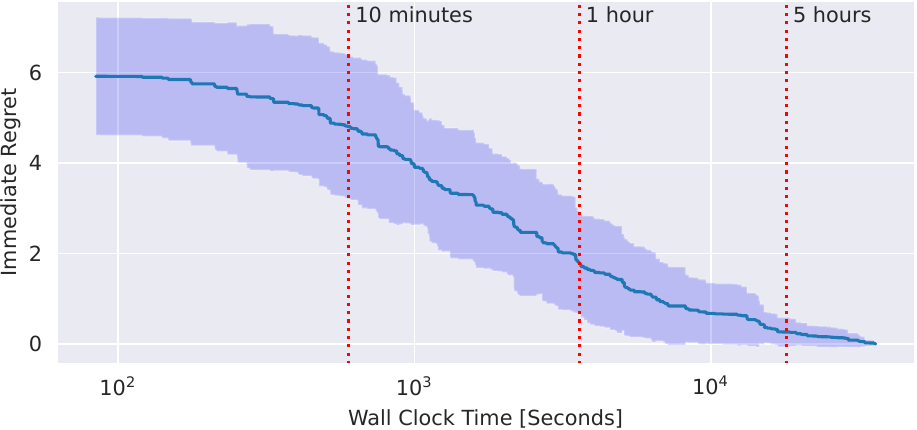}}
    \caption{Visualization of the mean performance, aggregated over all datasets and iterations, with standard deviations plotted over time. Displayed is the immediate regret, \ie the performance difference to the best solution, of the best so-far found configuration as a function over wall clock time.}
    \label{fig:performance-over-time}
\end{figure}

\paragraph{Results}
The results of all experiments are displayed in Table~\ref{tab:experiment-results}. It is apparent, that \name{AutoRUL} is able to outperform the current state-of-the-art models on the FD001, FD003, and Filtration datasets significantly according to Wilcoxon signed-rank test with \(\alpha = 0.05\). For the PHME'20, PRONOSTIA, and PHM'08 dataset, \name{AutoRUL} achieved the best average performance but at least one, but different, manual approaches were not significantly worse for each dataset. Similarly, on the FD002 the transformers approach proposed by \cite{Mo2021a} achieved the best result with \name{AutoML} being not significantly worse. Only on the FD004 dataset, the \ac{LSTM} approach proposed by \cite{Zheng2017} was able to significantly outperform all other approaches with \name{AutoRUL} in third place. In summary, \name{AutoRUL} outperformed the hand-crafted models in two cases, achieved state-of-the-art performance four times and was outperformed only once. Similarly, \name{AutoRUL} was also able to outperform \name{AutoCoevoRUL} which achieved only mediocre results.

\paragraph{Anytime Performance}
Next, we take a look at the \textit{anytime} performance of the optimizer, \ie how the performance of the best found solution changes over time. The performance of the optimizer after \(t\) seconds is measured using the immediate regret
\begin{equation*}
    r_t = \left|\mathcal{L}\left(h_{\vec{\lambda}^*_t}, \mathcal{D}_\mathrm{test}\right) - \mathcal{L}\left(h_{\vec{\lambda}^*}, \mathcal{D}_\mathrm{test}\right) \right|
\end{equation*}
with \(\vec{\lambda}^*\) being the best overall found solution and \(\vec{\lambda}^*_t\) the best solution found after \(t\) seconds. Fig.~\ref{fig:performance-over-time} visualizes the improvements during the optimization averaged over all datasets and repetitions. Even though improvements become smaller over time, even after multiple hours of search better configurations are still discovered. Yet, even with only five hours of optimization duration similar performances can be obtained. Further experiment results are also available in the online appendix.

\section{Conclusion \& Limitation}
\label{sec:conclusion}
The experiments have shown the viability of applying \ac{AutoML} for \ac{RUL} predictions. \name{AutoRUL} was able to generate at least competitive results on a wide variety of datasets. No other state-of-the-art model was able to generalize similarly over all datasets. This does not come as a surprise as all methods were tailored to specific datasets and not intended to be general-purpose \ac{RUL} predictors. This further proves that one-size-fits-all models are hard to craft and it may be better to build simple models automatically for each individual dataset.

\name{AutoRUL} enables domain experts without knowledge of \ac{ML} to build data-driven \ac{RUL} predictions. Furthermore, even users with \ac{ML} expertise can benefit from \name{AutoRUL} by generating pipeline baseline methods automatically that can be later on enhanced by incorporating additional domain knowledge, as described in Section~\ref{sec:comparison}.

In summary, we were able to show that a throughout exploration of simple models via \ac{AutoML} is able to compete with complex hand-crafted solutions containing domain knowledge.

\name{AutoRUL} can only handle direct \ac{RUL} predictions, \ie the true \ac{RUL} has to be known and available, making it unsuited for indirect \ac{RUL} predictions. Furthermore, it currently has quite high demands regarding the data quality, \eg equidistant measurements. As only \textit{high quality} datasets were considered for the evaluation this limitation is not directly visible. Making \name{AutoRUL} applicable to arbitrary datasets requires significantly more work. This shortcoming could be potentially compensated by domain experts. Yet, the fixed search space still prevents \name{AutoRUL} from being a universal tool as it will always be possible to construct a dataset it cannot handle.

Finally, replicating the results from \ac{RUL} literature proved harder than expected. Many baseline methods did not provide source code or sufficient implementation details to recreate the proposed models. We believe that the \ac{RUL} community will benefit from our study with several implemented open-source baselines, paving the way for more standardized benchmarks and the availability of source code to ensure fair comparisons.

\section*{Acknowledgements}
Marc Z\"oller acknowledges funding by the German Federal Ministry for Economic Affairs and Climate Action in the project FabOS (project no. 01MK20010N). Marius Lindauer acknowledges funding by the European Union (ERC, ``ixAutoML'', grant no.101041029). Views and opinions expressed are however those of the author(s) only and do not necessarily reflect those of the European Union or the European Research Council Executive Agency. Neither the European Union nor the granting authority can be held responsible for them. Marco Huber acknowledges funding by the Baden-Wuerttemberg Ministry for Economic Affairs, Labour and Tourism in the project KI-Fortschrittszentrum ``Lernende Systeme und Kognitive Robotik'' (project no. 036-170017).

\includegraphics[height=5cm]{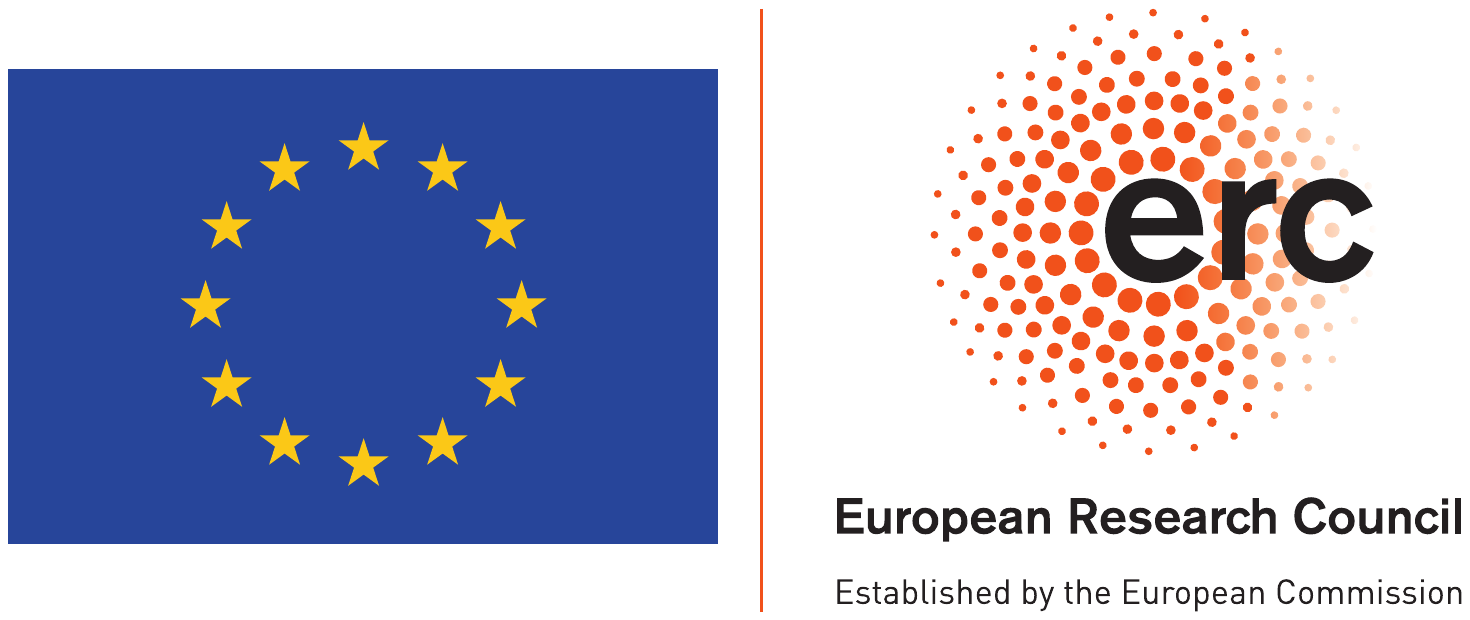}

\bibliography{library_marc,library_fabian}

\begin{thebibliography}{10}

\bibitem{Atamuradov.2017}
V.~Atamuradov, K.~Medjaher, P.~Dersin, B.~Lamoureux, and N.~Zerhouni.
\newblock Prognostics and health management for maintenance practitioners - review, implementation and tools evaluation.
\newblock {\em IJPHM}, 8(3), 2017.

\bibitem{Zio.2022}
E.~Zio.
\newblock Prognostics and health management (phm): Where are we and where do we (need to) go in theory and practice.
\newblock {\em Rel. Eng.}, 218, 2022.

\bibitem{Kordestani.2021}
M.~Kordestani, M.~Saif, M.~E. Orchard, R.~Razavi-Far, and K.~Khorasani.
\newblock Failure prognosis and applications---a survey of recent literature.
\newblock {\em IEEE Transactions on Reliability}, 70(2), 2021.

\bibitem{Mauthe2022}
F.~Mauthe, M.~Braig, and P.~Zeiler.
\newblock Performance evaluation of neural network architectures on time series condition data for remaining useful life prognosis under defined operating conditions.
\newblock In {\em ESREL}, 2022.

\bibitem{Hutter2018}
F.~Hutter, L.~Kotthoff, and J.~Vanschoren.
\newblock {\em Automated Machine Learning: Methods, Systems, Challenges}.
\newblock Springer, 2018.

\bibitem{Goebel.2008}
K.~Goebel, B.~Saha, A.~Saxena, N.~Mct, and N.~Riacs.
\newblock A comparison of three data-driven techniques for prognostics.
\newblock In {\em MFPT}, 2008.

\bibitem{JundaZhu.2014}
{J. Zhu}, {T. Nostrand}, {C. Spiegel}, and {B. Morton}.
\newblock Survey of condition indicators for condition monitoring systems.
\newblock {\em PHM}, 6(1), 2014.

\bibitem{Wang.2018}
Dong Wang, Kwok-Leung Tsui, and Qiang Miao.
\newblock Prognostics and health management: A review of vibration based bearing and gear health indicators.
\newblock {\em IEEE Access}, 6, 2018.

\bibitem{Christ2016}
M.~Christ, A.~Kempa-Liehr, and M.~Feindt.
\newblock Distributed and parallel time series feature extraction for industrial big data applications.
\newblock {\em arXiv preprint arXiv: 1610.07717}, 10 2016.

\bibitem{Lu.2019}
Yizhou Lu and Aris Christou.
\newblock Prognostics of igbt modules based on the approach of particle filtering.
\newblock {\em Microelectronics Reliability}, 92, 2019.

\bibitem{GarciaNieto.2015}
P.~J. Nieto, E.~Garc{\'i}a-Gonzalo, F.~{Lasheras}, and F.~J. de~{Cos Juez}.
\newblock Hybrid pso--svm-based method for forecasting of the remaining useful life for aircraft engines and evaluation of its reliability.
\newblock {\em Rel. Eng.}, 138, 2015.

\bibitem{Kuersat2020}
Ince Kürsat, Engin Sirkeci, and Yakup Genç.
\newblock Remaining useful life prediction for experimental filtration system.
\newblock In {\em PHME}, 2020.

\bibitem{Mo2021a}
Yu~Mo, Qianhui Wu, Xiu Li, and Biqing Huang.
\newblock Remaining useful life estimation via transformer encoder enhanced by a gated convolutional unit.
\newblock {\em Journal of Intelligent Manufacturing}, 32:1997--2006, 2021.

\bibitem{Shi.2021}
Z.~Shi and A.~Chehade.
\newblock A dual-lstm framework combining change point detection and remaining useful life prediction.
\newblock {\em Rel. Eng.}, 205, 2021.

\bibitem{Wang.2021}
C.~Wang, W.~Jiang, X.~Yang, and S.~Zhang.
\newblock Rul prediction of rolling bearings based on a dcae and cnn.
\newblock {\em Applied Sciences}, 11(23), 2021.

\bibitem{Braig.2023}
M.~Braig and P.~Zeiler.
\newblock Using data from similar systems for data-driven condition diagnosis and prognosis of engineering systems: A review and an outline of future research challenges.
\newblock {\em IEEE Access}, 11, 2023.

\bibitem{Hagmeyer.2022}
S.~Hagmeyer, P.~Zeiler, and M.~Huber.
\newblock On the integration of fundamental knowledge about degradation processes into data-driven diagnostics and prognostics using theory-guided data science.
\newblock {\em PHME}, 7(1), 2022.

\bibitem{Tornede2020}
T.~Tornede, A.~Tornede, M.~Wever, F.~Mohr, and E.~Hüllermeier.
\newblock Automl for predictive maintenance: One tool to rul them all.
\newblock In {\em IoT Streams 2020}, 2020.

\bibitem{Tornede2021}
T.~Tornede, A.~Tornede, M.~Wever, and E.~Hüllermeier.
\newblock Coevolution of remaining useful lifetime estimation pipelines for automated predictive maintenance.
\newblock In {\em GECCO}, 2021.

\bibitem{Singh2022}
M.~Singh, S.~Bansal, Vandana, B.~K. Panigrahi, and A.~Garg.
\newblock A genetic algorithm and rnn-lstm model for remaining battery capacity prediction.
\newblock {\em J. Comput. Inf. Sci. Eng.}, 22, 2022.

\bibitem{Zhou2020}
Y.~Zhou, Y.~Gao, Y.~Huang, M.~Hefenbrock, T.~Riedel, and M.~Beigl.
\newblock Automatic remaining useful life estimation framework with embedded convolutional lstm as the backbone.
\newblock In {\em ECML-PKDD}, 2020.

\bibitem{Lindauer2022}
M.~Lindauer, K.~Eggensperger, M.~Feurer, A.~Biedenkapp, D.~Deng, C.~Benjamins, R.~Sass, and F.~Hutter.
\newblock Smac3: A versatile bayesian optimization package for hyperparameter optimization.
\newblock {\em JMLR}, 23, 2022.

\bibitem{Li2018}
L.~Li, K.~Jamieson, G.~DeSalvo, and A.~Talwalkar.
\newblock Hyperband: A novel bandit-based approach to hyperparameter optimization.
\newblock {\em JMLR}, 18, 2018.

\bibitem{Caruana2004}
Rich Caruana, Alexandru Niculescu-Mizil, Geoff Crew, and Alex Ksikes.
\newblock Ensemble selection from libraries of models.
\newblock {\em ICML}, 2004.

\bibitem{Zheng2017}
S.~Zheng, K.~Ristovski, A.~Farahat, and C.~Gupta.
\newblock Long short-term memory network for remaining useful life estimation.
\newblock In {\em ICPHM}, 2017.

\bibitem{Li2018b}
X.~Li, Q.~Ding, and J.~Sun.
\newblock Remaining useful life estimation in prognos-tics using deep convolution neural networks.
\newblock {\em Rel. Eng.}, 172, 2017.

\bibitem{Saxena2008}
A.~Saxena, K.~Goebel, D.~Simon, and N.~Eklund.
\newblock Damage propagation modeling for aircraft engine run-to-failure simulation.
\newblock In {\em ICPHM}, 2008.

\bibitem{Giordano2020}
Danilo Giordano and Daniel Gagar.
\newblock Phme 2020 data challenge, 2020.

\bibitem{Nectoux2012}
P.~Nectoux, R.~Gouriveau, K.~Medjaher, E.~Ramasso, B.~Chebel-Morello, N.~Zerhouni, and C.~Varnier.
\newblock Pronostia: An experimental platform for bearings accelerated degradation tests.
\newblock In {\em ICPHM}, 2012.

\end{thebibliography}
\bibliographystyle{unsrt}

\end{document}


\maketitle

\vspace{-10ex}
\section{Search Space Description}
The search space \(\Lambda\) of \name{AutoRUL} allows the creation of \(624\) unique pipelines, which can be further configured by a total of \(168\) hyperparameters. In the following tables, more details regarding the available algorithms and hyperparameters are provided. For each algorithm, the total number of hyperparameters (\(\#\lambda\)), the number of categorical (cat) and numerical (num) hyperparameters, is given. Numbers in parentheses denote conditional hyperparameters. The total number of hyperparameters does not add up to the reported 168 hyperparameters as hyperparameters of some algorithms, like, for example, TS Fresh or window generation, are included twice (once for tabular regression and once for sequence-to-sequence regression using neural networks). Components highlighted in italic are not directly included in the pipeline but influence the fitting procedure of the pipeline.

\begin{table}[ht!]
  \mbox{}\hfill
  \begin{minipage}[t]{.48\linewidth}
    \caption{Preprocessing Algorithms}
    \centering
    \begin{tabular}{@{} l c c c @{}}
        \toprule
        Name                    & \(\#\lambda\) & cat & num \\
        \midrule
        Imputation              & 1 & 1 (0) & 0 (0) \\
        Exponential Smoothing   & 2 & 0 (0) & 2 (0) \\
        Robust Scaler           & 2 & 0 (0) & 2 (0) \\
        Normalizer              & 0 & 0 (0) & 0 (0) \\
        Min Max Scaling         & 0 & 0 (0) & 0 (0) \\
        Standardizer            & 0 & 0 (0) & 0 (0) \\
        \bottomrule
    \end{tabular}
  \end{minipage}
  \hfill
  \begin{minipage}[t]{.48\linewidth}
    \caption{Feature Engineering Algorithms}
    \centering
    \begin{tabular}{@{} l c c c @{}}
        \toprule
        Name                & \(\#\lambda\) & cat & num \\
        \midrule
        Window Generation    & 2  & 0 (0)  & 2 (0) \\
        
        Flattening          & 0  & 0 (0)  & 0 (0) \\
        TS Fresh            & 43 & 43 (0) & 0 (0) \\
        PCA                 & 2  & 1 (0)  & 1 (0) \\
        Select Percentile   & 2  & 1 (0)  & 1 (0) \\
        Select Rates        & 3  & 2 (0)  & 1 (0) \\
        \bottomrule
    \end{tabular}
  \end{minipage}
  \mbox{}
\end{table}

\begin{table}[ht!]
  \mbox{}\hfill
  \begin{minipage}[t]{.48\linewidth}
    \caption{Tabular Regression Algorithms}
    \centering
    \begin{tabular}{@{} l c c c @{}}
        \toprule
        Name                & \(\#\lambda\) & cat & num \\
        \midrule
        Extra Trees         & 5 & 2 (0) & 3 (0) \\
        Gradient Boosting   & 6 & 0 (0) & 6 (0) \\
        MLP                 & 6 & 2 (0) & 4 (1) \\
        Passive Aggressive  & 4 & 2 (0) & 2 (0) \\
        Random Forest       & 3 & 1 (0) & 2 (0) \\
        SGD                 & 6 & 2 (0) & 4 (1) \\
        \bottomrule
    \end{tabular}
  \end{minipage}
  \hfill
  \begin{minipage}[t]{.48\linewidth}
    \caption{Sequence Regression Algorithms}
    \centering
    \begin{tabular}{@{} l c c c @{}}
        \toprule
        Name        & \(\#\lambda\) & cat & num \\
        \midrule
        \textit{Optimizer}  & 4 & 0 (0) & 4 (0) \\
        \textit{Trainer}    & 2 & 0 (0) & 2 (0) \\
        CNN                 & 5 & 1 (0) & 4 (1) \\
        GRU                 & 4 & 1 (0) & 3 (1) \\
        LSTM                & 4 & 1 (0) & 3 (1) \\
        TCN                 & 5 & 1 (0) & 4 (1) \\
        \bottomrule
    \end{tabular}
  \end{minipage}
  \mbox{}
\end{table}

\section{Detailed Experimental Results}
\begin{figure}[ht!]
    \centering
    \begin{subfigure}[t]{0.49\textwidth}
        \centering
        \includegraphics[width=\textwidth]{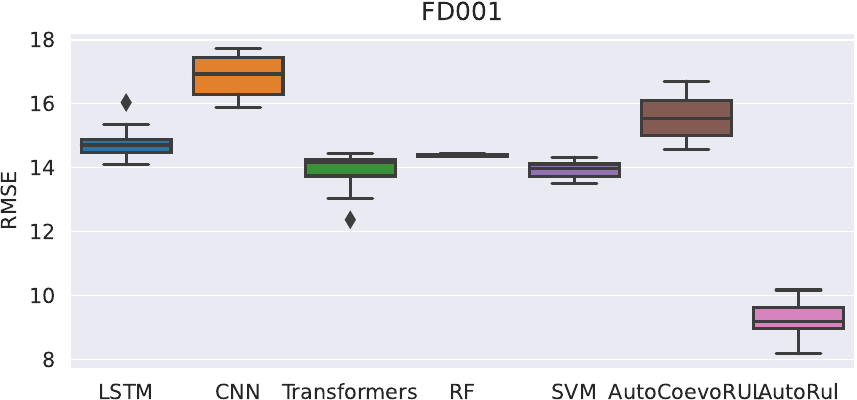}
    \end{subfigure}
    \hfill
    \begin{subfigure}[t]{0.49\textwidth}
        \centering
        \includegraphics[width=\textwidth]{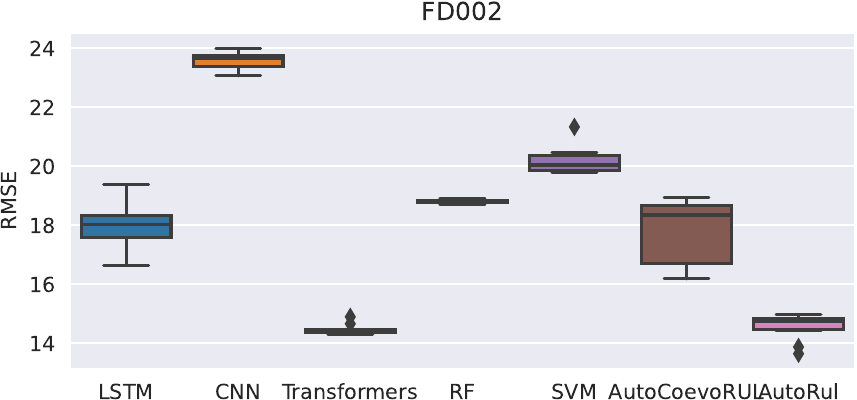}
    \end{subfigure}

    \vspace{4ex}

    \begin{subfigure}[t]{0.49\textwidth}
        \centering
        \includegraphics[width=\textwidth]{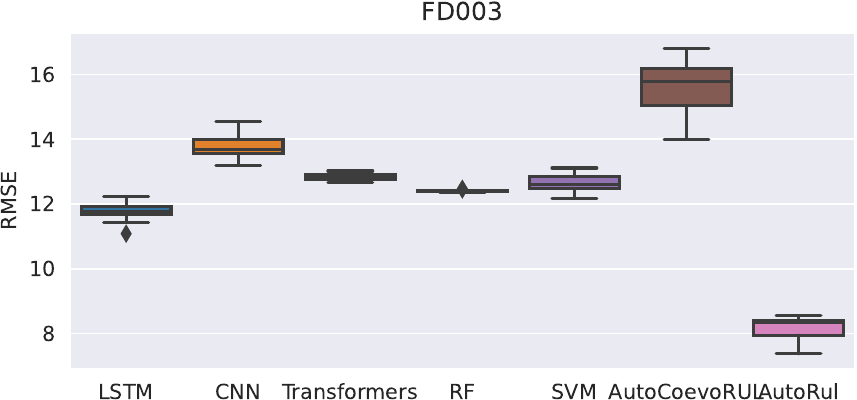}
    \end{subfigure}
    \hfill
    \begin{subfigure}[t]{0.49\textwidth}
        \centering
        \includegraphics[width=\textwidth]{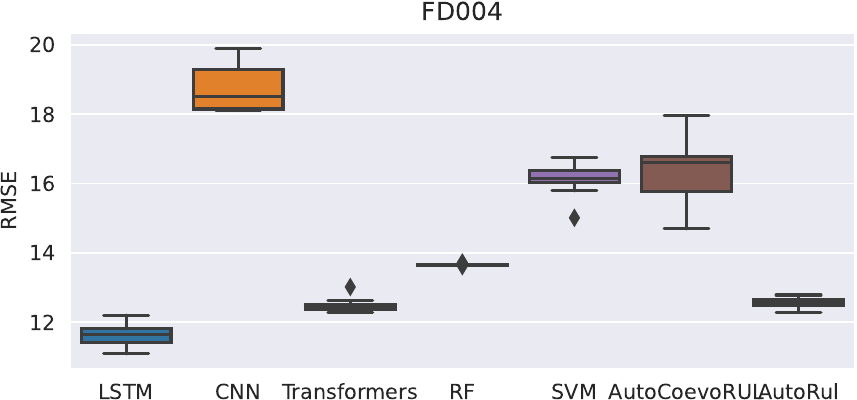}
    \end{subfigure}

    \vspace{4ex}

    \begin{subfigure}[t]{0.49\textwidth}
        \centering
        \includegraphics[width=\textwidth]{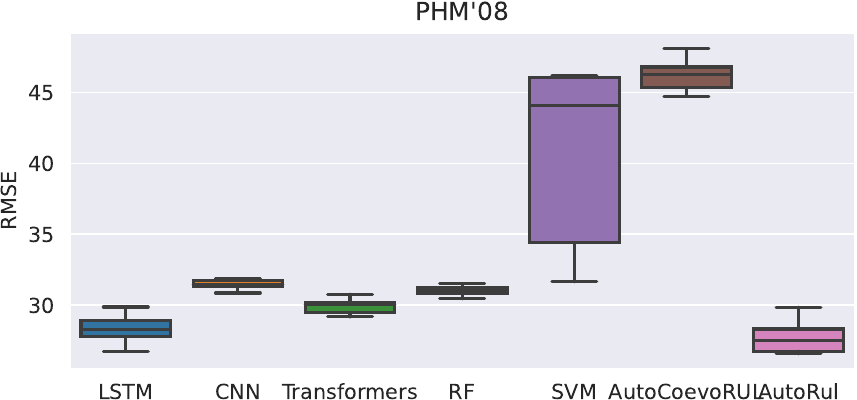}
    \end{subfigure}
    \hfill
    \begin{subfigure}[t]{0.49\textwidth}
        \centering
        \includegraphics[width=\textwidth]{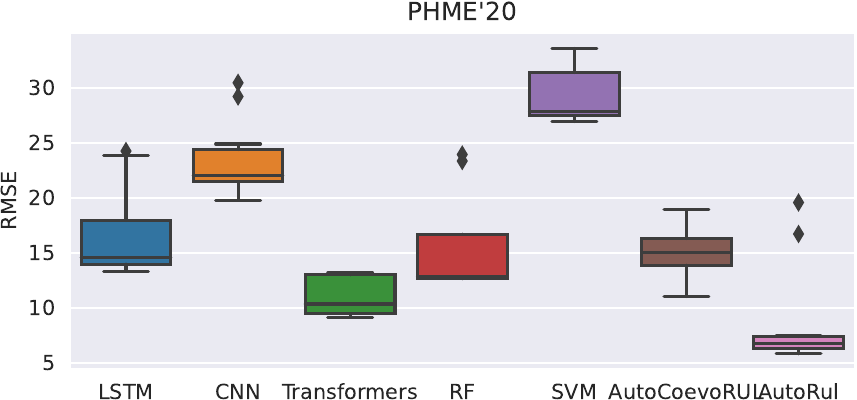}
    \end{subfigure}

    \vspace{4ex}

    \begin{subfigure}[t]{0.49\textwidth}
        \centering
        \includegraphics[width=\textwidth]{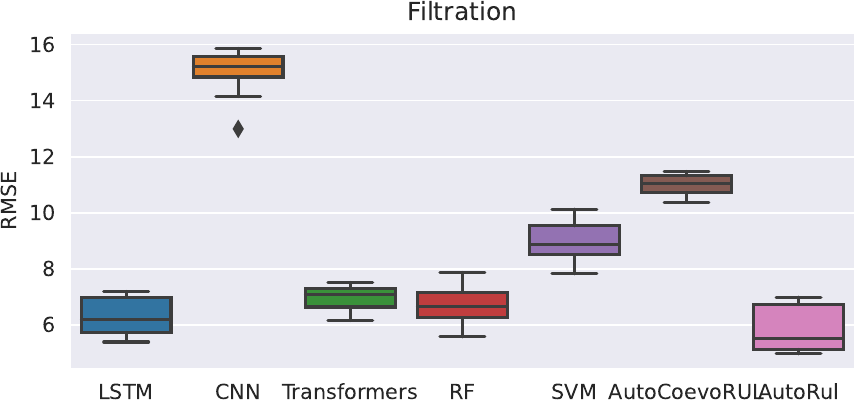}
    \end{subfigure}
    \hfill
    \begin{subfigure}[t]{0.49\textwidth}
        \centering
        \includegraphics[width=\textwidth]{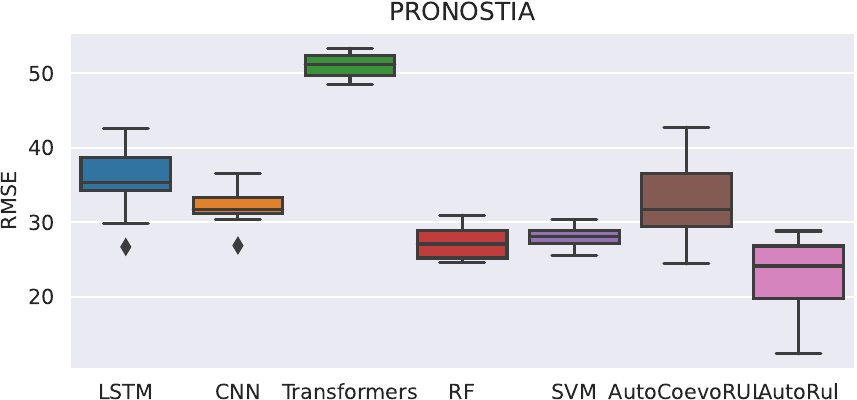}
    \end{subfigure}
    
    \caption{Performance visualizations of all tested RUL methods on all benchmark datasets.}
    \label{fig:final-performance}
\end{figure}

Besides the results reported in the manuscript, we want to provide further insights into the generated pipelines. Figure~\ref{fig:final-performance} contains visualizations of the final performances reported in \textit{Table~1} of the manuscript per benchmark dataset.

\pagebreak

Next, we take a closer look at the behaviour of \name{AutoRUL} during the optimization. Table~\ref{tab:configurations} contains statistics about the number of evaluated configurations for each dataset. Reported are the total number of evaluated configurations, the number of successful evaluations, the number of failed evaluations (for example due to an exception during model fitting), the number of configurations where fitting was canceled after five minutes, and the number of configurations that were trained on the complete training data (no multi-fidelity approximation). In general, the vast majority of the evaluated configurations was fitted successfully. Less than 3\% of the evaluated configurations were aborted after the configured timeout of five minutes, indicating that the limit was not selected to aggressively. Even though a large number of different configurations was evaluated for each dataset, only roughly 5\% of the configurations were evaluated on the full budget, i.e.,~fully fitted until convergence.

\begin{table}[ht!]
    \caption{Statistics about the evaluated configurations. Results are averaged over ten repetitions.}
    \label{tab:configurations}
    \centering
    \begin{tabular}{@{} l c c c c c c @{}}
        \toprule
        Dataset     & \# Configurations & \# Success & \# Failed & \# Timeout & \# Full Budget \\
        \midrule

        FD001       & \(761.8 \pm 144.48\) & \(739.1 \pm 135.70\) & ~\(8.5 \pm 9.92\)   & \(14.2 \pm 6.71\)~  & \(41.3 \pm 7.93\)~\\
        FD002       & \(932.5 \pm 382.89\) & \(897.2 \pm 391.24\) & ~\(9.1 \pm 7.71\)   & \(26.2 \pm 14.28\)  & \(50.2 \pm 21.61\)\\
        FD003       & \(648.8 \pm 62.75\)~ & \(631.9 \pm 65.17\)~ & ~\(3.3 \pm 1.55\)   & \(13.6 \pm 5.64\)~  & \(35.4 \pm 3.32\)~\\
        FD004       & \(991.9 \pm 462.97\) & \(923.0 \pm 356.89\) & \(46.7 \pm 5.83\)   & \(22.2 \pm 5.83\)~  & \(53.2 \pm 26.46\)\\
        PHM'08      & \(355.0 \pm 26.65\)~ & \(323.1 \pm 25.04\)~ & \(28.5 \pm 0.92\)   & \(28.5 \pm 4.34\)~  & \(18.7 \pm 1.55\)~\\
        PHME'20     & \(886.9 \pm 125.11\) & \(875.0 \pm 122.06\) & ~\(8.0 \pm 2.47\)   & ~\(8.0 \pm 4.90\)~  & \(48.4 \pm 6.48\)~\\
        Filtration  & \(818.8 \pm 134.86\) & \(799.0 \pm 137.68\) & ~\(5.7 \pm 3.29\)   & \(14.1 \pm 10.09\)  & \(44.7 \pm 7.87\)~\\
        PRONOSTIA   & \(471.5 \pm 79.13\)~ & \(354.7 \pm 76.85\)  & \(86.8 \pm 12.84\)  & \(30.0 \pm 9.15\)~  & \(25.3 \pm 4.50\)~\\
        \bottomrule
    \end{tabular}
\end{table}

Finally, we take a closer look at the pipelines constructed by \name{AutoRUL}. Table~\ref{tab:pipelines} provides an overview of the constructed ensembles and the pipelines in them for each of the benchmark datasets. The \textit{ensemble size} column shows the average number of pipelines in the constructed ensemble. It is apparent that the option to build ensembles is used for all datasets but the maximum ensemble size of ten pipelines is not consistently reached. Regarding the used template, both sequence-to-sequence regression and tabular regression are used. Depending on the dataset, either one of the two options can be basically used exclusively or both options can be mixed together. Similarly, for feature engineering all three available algorithms are extensively used with TS Fresh being used the most. Finally, for most datasets a specific type of regression algorithm is used significantly more often than others. Only for the PHM'08 dataset four different algorithm types are used roughly equally often. Random forests and TCNs are by far the most used regression algorithms. In contrast, SGD and extra trees are not included in any of the final ensembles and can probably pruned from the search space. 

\begin{table}[ht!]
    \caption{Overview of constructed pipelines for each benchmark dataset. Numbers in parentheses represent the fraction of pipelines using the according component. The sum of fractions within one cell can be unequal to 1.00 due to rounding errors.}
    \label{tab:pipelines}
    \centering
    \setlength{\extrarowheight}{1.5ex}
    \setlength{\tabcolsep}{1.2em}
    \begin{tabular}{@{} l l @{\hskip 10pt} l l l @{}}
        \toprule
        Dataset     & Ensemble Size &   Template    & Feat. Eng. & Regressor \\  
        \midrule
        FD001       & \(8.6 \pm 0.92\)           &   \makecell[lt]{seq2seq (1.0)} & \makecell[lt]{Flatten (0.35) \\ Limited (0.01) \\ TS Fresh (0.64)} & \makecell[lt]{CNN (0.20) \hspace{1.5em} GRU (0.03) \\ LSTM (0.02) \hspace{1em} TCN (0.74)} \\
        FD002 & \(6.0 \pm 1.84\) & \makecell[lt]{seq2seq (0.15) \\ Tabular (0.85)} & \makecell[lt]{Flatten (0.12) \\ Limited (0.67) \\ TS Fresh (0.22)} & \makecell[lt]{Gradient Boosting (0.13) \\ Random Forest (0.72) \\ GRU (0.03) \hspace{1.7em} LSTM (0.02) \\ TCN (0.10)} \\
        FD003 & \(7.8 \pm 1.17\) & \makecell[lt]{seq2seq (0.95) \\ Tabular (0.05)} & \makecell[lt]{Flatten (0.31) \\ Limited (0.08) \\ TS Fresh (0.61)} & \makecell[lt]{Gradient Boosting (0.01) \\ Random Forest (0.04) \\ CNN (0.10) \hspace{1.5em} GRU (0.01) \\ LSTM (0.03) \hspace{1em} TCN (0.81)} \\
        FD004 & \(5.9 \pm 1.22\) & \makecell[lt]{seq2seq (0.15) \\ Tabular (0.85)} & \makecell[lt]{Flatten (0.16) \\ Limited (0.36) \\ TS Fresh (0.47)} & \makecell[lt]{Gradient Boosting (0.03) \\ Random Forest (0.81) \\ CNN (0.07) \\ TCN (0.08)} \\
        PHM'08 & \(5.8 \pm 1.83\) & \makecell[lt]{seq2seq (0.74) \\ Tabular (0.26)} & \makecell[lt]{Flatten (0.18) \\ Limited (0.54) \\ TS Fresh (0.30)} & \makecell[lt]{Passive Aggressive (0.02) \\ Random Forest (0.24) \\ CNN (0.33) \hspace{1.5em} GRU (0.16) \\ LSTM (0.05) \hspace{1em} TCN (0.21)} \\
        PHM'20 & \(7.2 \pm 1.17\) & \makecell[lt]{seq2seq (0.99) \\ Tabular (0.01)} & \makecell[lt]{Flatten (0.05) \\ Limited (0.04) \\ TS Fresh (0.90)} & \makecell[lt]{Passive Aggressive (0.01) \\ CNN (0.01) \hspace{1.5em} GRU (0.36) \\ LSTM (0.33) \hspace{1em} TCN (0.28)} \\
        Filtration & \(5.8 \pm 1.83\) & \makecell[lt]{seq2seq (0.14) \\ Tabular (0.86)} & \makecell[lt]{Flatten (0.10) \\ TS Fresh (0.90)} & \makecell[lt]{Gradient Boosting (0.03) \\ Random Forest (0.83) \\ GRU (0.07) \hspace{1.7em}  LSTM (0.03)\\ TCN (0.03)} \\
        PRONOSTIA & \(4.5 \pm 1.12\) & \makecell[lt]{Tabular (1.00)} & \makecell[lt]{Flatten (0.24) \\ Limited (0.27) \\ TS Fresh (0.49)} & \makecell[lt]{Gradient Boosting (0.16) \\ MLP (0.18) \\ Passive Aggressive (0.18) \\ Random Forest (0.49)} \\
        \bottomrule
    \end{tabular}

\end{table}